\documentclass[letterpaper, 10 pt, conference]{ieeeconf}  

\IEEEoverridecommandlockouts                              

\usepackage{graphics} 
\usepackage{epsfig} 
\usepackage{mathptmx} 
\DeclareMathAlphabet{\mathcal}{OMS}{cmsy}{m}{n}
\usepackage{times} 
\usepackage{amsmath} 
\usepackage{amssymb}  
\usepackage{float}
 \usepackage{bm}
 \usepackage{xspace}
\usepackage{graphicx}
\graphicspath{./figures/}
\usepackage{adjustbox}
\usepackage{svg}

\usepackage{tikz}
\usepackage{amsmath}
\usepackage{amssymb}
\usepackage{bm}
\usepackage{tabularx}
\usepackage{booktabs}
\usepackage{array}

\usepackage{caption}
\usepackage{subcaption}
\usepackage{algorithm}
\usepackage{algpseudocode}
\usepackage{upgreek}
\usepackage{hyperref}
\usepackage{makecell}

\usepackage{url}
\usepackage{xurl}
\usepackage{color}
\usepackage{tikz}

\usepackage{pgfplots}
\usepackage{multirow}
\usepackage{colortbl}
\usepackage{wasysym}

\usepackage{todonotes}

\usepackage{listings}

\title{\LARGE \bf
TACTIC: Temporal and Context-Aware LLM Tactical Planning for Roadside LiDAR Attacks
}

\author{
}
\author{
    Yiming Gao$^{1}$\thanks{$^{1}$Y. Gao is with the Department of Electrical Engineering and Computer Science, University of Michigan, Ann Arbor, MI, USA. \texttt{gdorothy@umich.edu}.} and
    Shaocheng Luo$^{2}$*\thanks{$^{2}$S. Luo is with the Department of Electrical and Computer Engineering, Duke University, Durham, NC, USA.}
    \thanks{*Corresponding Author: S. Luo, \texttt{shaocheng.luo@duke.edu}.}
}

\begin{document}

\maketitle
\thispagestyle{empty}
\pagestyle{empty}


\begin{abstract}
Physical LiDAR attacks are often evaluated using fixed primitives and manually selected parameters, despite their strong dependence on surrounding traffic. We present \textsc{TACTIC}, a scene-aware framework that uses a multimodal large language model (MLLM) to coordinate state-adaptive roadside LiDAR attacks. Under a gray-box threat model, \textsc{TACTIC} relies only on an attacker-operated roadside perception stack, without accessing the victim LiDAR's native point clouds or internal processing. Local perception provides metric vehicle states, while the MLLM combines these measurements with roadside imagery to infer relational traffic context and construct a semantic scene graph. Based on this representation, \textsc{TACTIC} selects and configures two complementary primitives: \emph{push-away}, which shifts the perceived range of a lead vehicle, and \emph{phantom-obstacle braking}, which triggers emergency braking through obstacle injection. Measured traffic states and empirically calibrated constraints ground the generated tactics in physically feasible operating regions. To accommodate MLLM latency, \textsc{TACTIC} overlaps reasoning and execution asynchronously while high-rate local perception detects scene changes and triggers replanning. Across 280 randomized CARLA trials, the full policy achieves a 100\% collision rate, compared with 35\% for a fixed rule, 60\% for random selection, and 75\% for a restricted LLM using mode selection with default parameters. Joint physical-and-image input achieves 100\% success, versus 65\% with physical measurements alone and 75\% with imagery alone, while asynchronous $\Delta$ refresh reduces scene-mutation response from 7.4\,s to 2.0\,s. These results show that scene-dependent tactical planning can expose context-sensitive LiDAR failure modes that fixed attack policies may miss.
\end{abstract}

\section{Introduction}
\label{sec:introduction}

LiDAR is widely used in autonomous driving but remains vulnerable to physical attacks that suppress genuine returns, inject phantom objects, and induce unsafe vehicle behavior \cite{cao2019adversarial,jin2023pla,zhang2025sok}. Studying these attacks in controlled environments is essential for identifying safety-critical failure modes, evaluating countermeasures, and improving protection for passengers and other road users.

Existing physical LiDAR attacks mainly demonstrate individual attack primitives. A target may be removed through spoofing \cite{cao2023you}, while a nonexistent obstacle may be introduced through point injection \cite{jin2023pla}. However, these attacks are typically evaluated using manually defined rules and fixed parameters in simplified traffic settings, such as straight-road driving \cite{sato2025realism}, predefined turning scenarios \cite{zhang2025ghost}, or pre-assumed vehicle maneuvers \cite{song2025gradient}. Such evaluations establish whether an individual primitive can work, but provide limited insight into how attack effectiveness changes with surrounding traffic scenes.

We focus on vehicle-to-vehicle interactions because they directly expose these dynamic dependencies and are central to autonomous-driving safety. Unlike collisions with walls or road dividers, which are dominated by static geometry and may additionally depend on map availability, localization, or lateral control, interactions among vehicles depend on continuously evolving traffic relationships. A nearby vehicle may be the target's leader or follower in the same lane, or an unrelated participant in an adjacent lane; the same perturbation can therefore produce very different outcomes. Physical attack execution is consequently a \emph{scene-dependent planning} problem that requires reasoning over traffic topology, spacing, relative motion, acceleration, and time-to-collision (TTC), rather than applying a fixed primitive or parameter set.

\begin{figure*}[t]
\centering
\includegraphics[width=\textwidth]{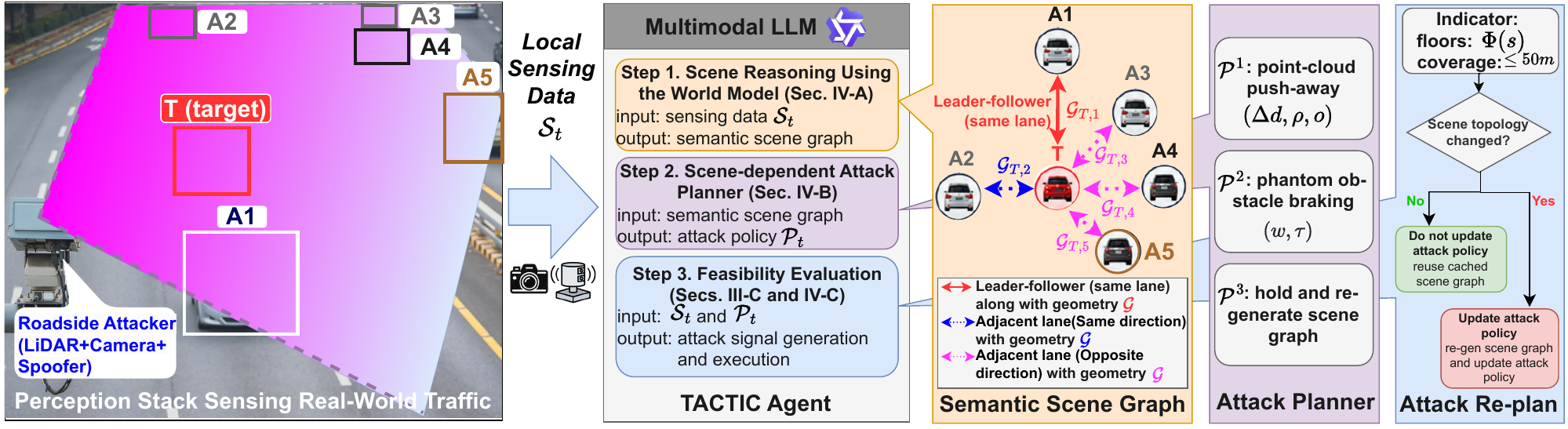}
\caption{\textsc{TACTIC} overview. The attacker-operated roadside perception stack observes local traffic with co-located LiDAR and camera and provides sensing state $\mathcal{S}$ to the multimodal LLM agent. The agent constructs a semantic scene graph, selects a feasible tactic $\mathcal{P}$, and updates the policy when significant scene changes occur. Cached topology and asynchronous reasoning reduce repeated scene generation and replanning latency.}
\vspace{-10pt}
\label{fig:overview}
\end{figure*}

We address this gap with \textsc{TACTIC}, a temporal and context-aware framework for state-adaptive, LLM-orchestrated roadside LiDAR attacks, illustrated in Fig.~\ref{fig:overview}. Under a gray-box threat model, the roadside attacker independently observes traffic without accessing the victim LiDAR's native point clouds, filtering parameters, or internal processing. \textsc{TACTIC} combines object-level local perception with roadside imagery and uses a multimodal large language model (MLLM) for relational scene reasoning and context-dependent tactical planning.

Specifically, local perception detects and tracks individual vehicles and provides measurable physical states such as position, velocity, spacing, and TTC, while the MLLM jointly considers these measurements and visual road context to infer same-lane, adjacent-lane, and leader--follower relationships. These relations form a semantic \emph{scene graph} for tactical decision making. Based on this graph, \textsc{TACTIC} selects and configures two complementary collision primitives: a \emph{push-away attack}, which shifts the preceding vehicle's perceived range, and a \emph{phantom-obstacle braking attack}, which triggers emergency braking through obstacle injection. Because the primitives exploit different vehicle relationships and feasibility regions, their effectiveness depends on the current scene.

Introducing an LLM into the loop creates two additional challenges: generated tactics may violate physical feasibility, and LLM inference may lag behind rapidly changing traffic. \textsc{TACTIC} addresses the former using measured vehicle states, calibrated operating bounds, and deterministic verification, leaving the MLLM to focus on relational scene understanding and tactical selection. To address latency, reasoning and execution overlap asynchronously, while high-rate local perception detects scene changes and triggers replanning only when needed.

We evaluate \textsc{TACTIC} in CARLA using randomized multi-vehicle scenarios, a realistic driver model, and a two-stage automatic emergency braking controller. Across 280 primary CARLA trials spanning feasibility, policy, perception, timing, and benign-control evaluations, the full LLM policy achieves a 100\% collision rate, compared with 35\% for a fixed rule and 60\% for random selection. A restricted LLM that selects the attack mode but uses default parameters reaches 75\%, showing that joint mode and parameter adaptation provides additional benefit. Joint physical and visual input sustains 100\% success, whereas physical-only and image-only inputs drop to 65\% and 75\%, respectively. Physical calibration reveals a sharp push-away feasibility transition near 10.5--11\,m, while asynchronous replanning reduces scene-change response latency from 7.4\,s to 2.0\,s. The project website\footnote{https://sites.google.com/umich.edu/tactic/home} is accessible, and the source code\footnote{https://github.com/gdorothy/TACTIC-Roadside-LiDAR-Attack} is released.

The main contributions are as follows:
\begin{itemize}
    \item We formulate physical LiDAR attack execution as a \emph{scene-dependent tactical planning problem}, focusing on dynamic vehicle-to-vehicle interactions whose feasibility depends on evolving relational traffic context.

    \item We develop \textsc{TACTIC}, which combines object-level local perception and roadside imagery with multimodal LLM reasoning to construct semantic scene graphs and select complementary physical attack primitives and parameters.

    \item We ground LLM tactical decisions with measured vehicle states and calibrated physical constraints and introduce an asynchronous perception--LLM architecture for high-rate scene-change detection and replanning. Experiments demonstrate higher collision success than fixed and random policies and faster adaptation to changing scenes.
\end{itemize}

\section{Related Work}
\label{sec:relatedwork}

\subsection{Physical LiDAR Attacks on Autonomous Vehicles}

Physical LiDAR attacks largely follow two directions. \emph{Removal attacks} suppress genuine target returns through adversarial objects or physical spoofing \cite{cao2019adversarial,tu2020physically,cao2023you,sun2020robust}, whereas \emph{injection attacks} introduce nonexistent obstacles to manipulate downstream perception and control \cite{jin2023pla,nassi2020phantom}. Recent work improves operational realism by attacking moving targets at longer range \cite{sato2025realism} and extending physical manipulation to localization, odometry, and sensor fusion \cite{zhang2025ghost,song2025gradient}. A recent systematization further characterizes how sensor perturbations propagate through autonomous-driving pipelines \cite{zhang2025sok}.

Despite these advances, attack execution remains largely scenario-specific: individual primitives are commonly evaluated with predefined conditions and parameters tailored to particular road configurations or maneuvers \cite{sato2025realism,zhang2025ghost,song2025gradient}. How to select and configure physical attacks as multi-vehicle traffic evolves has received substantially less attention. \textsc{TACTIC} addresses this gap by formulating physical attack execution as a scene-dependent tactical planning problem.

\subsection{Context-Aware Attack Planning}

Context-dependent coordination of physical sensor attacks remains comparatively unexplored, with existing approaches typically executing individual attacks independently or using handcrafted rules that map predefined conditions to fixed actions. Attack-hardness analysis \cite{kim2024hardness} shows that physical attack feasibility depends strongly on situation-specific conditions, but identifies these conditions primarily through offline search. LLM-based adversarial scenario generation \cite{mei2025llmattacker} instead varies simulated driving scenarios rather than coordinating online physical attacks. Zhang et al.\ \cite{zhang2025sok} further identify joint perception--decision attacks and scene-aware hybrid attack chains as open directions.

\textsc{TACTIC} moves this reasoning online: measured traffic states and calibrated physical constraints define the feasible action space, while relational scene understanding guides context-dependent selection and parameterization of complementary physical attack primitives.

\subsection{LLM-Based Scene and Tactical Reasoning}

LLMs and multimodal LLMs (MLLMs) are increasingly used for high-level reasoning in autonomous driving. GPT-Driver reformulates motion planning as language modeling over structured driving states \cite{mao2023gpt}; LMDrive integrates multimodal sensor observations and language for closed-loop driving \cite{shao2024lmdrive}; and DriveLM introduces graph-structured visual reasoning across perception, prediction, and planning \cite{sima2024drivelm}. These studies demonstrate the potential of language models to reason over heterogeneous observations and relationships among traffic participants.

\textsc{TACTIC} extends this capability to physical-attack planning: local perception provides metric states, the MLLM infers relational context and selects tactics, while physical feasibility and execution remain deterministic. To reduce latency, \textsc{TACTIC} overlaps reasoning with execution following TypeFly's generate-while-execute principle \cite{chen2025typefly}.

\section{Scene-Dependent Attack Formulation}
\label{sec:formulation}

We focus on vehicle-to-vehicle interactions and stress tests because they require reasoning over dynamic traffic relationships, whereas collisions with walls or road dividers are dominated by static geometry and may additionally depend on map availability, localization, and lateral control. These factors can confound the study of physical LiDAR attacks and obscure the role of surrounding traffic. Our work provides a direct setting for studying how dynamic relationships, including leader--follower structure, spacing, and relative motion, affect attack feasibility.

This objective makes physical attack execution inherently
\emph{scene dependent}. As shown in Fig.~\ref{fig:overview}, a nearby vehicle may be the target's leader or follower in the same lane, or an unrelated participant in an adjacent lane. The same physical primitive can therefore succeed or fail depending on the current traffic topology and vehicle dynamics. We formulate this scene-dependent attack space below; Sec.~\ref{sec:tactic} then introduces \textsc{TACTIC} to reason over the scene and select among feasible attack policies.

\subsection{Threat Model and Attack Objective}
\label{sec:attack_obj}

We consider a roadside attacker equipped with LiDAR and an RGB camera, following a deployment setting similar to the Moving Vehicle Spoofing System in \cite{sato2025realism}. The roadside perception stack detects and tracks surrounding vehicles at 20 Hz. Vehicles follow an intelligent driver
model \cite{kesting2010enhanced,yao2026seidm} together with a two-stage AEB controller that applies partial and full braking at perceived TTC thresholds of 3.0\,s and 1.6\,s, respectively \cite{euroncap2024}.

We adopt a gray-box threat model in which the attacker interacts with the victim only through the physical sensing channel. The attacker independently observes the traffic scene and may know relevant sensor and vehicle-control configurations, but has no access to the victim's native point clouds, internal filtering states, or vehicle messages. A target vehicle $T$ is designated, and the objective is to induce a collision between $T$ and a neighboring traffic participant.

Let $\mathcal{S}_t$ denote the roadside observation at time $t$, comprising vehicle positions, velocities, accelerations, inter-vehicle spacing, TTC, and the corresponding roadside image. Because these object-level measurements do not by themselves encode how traffic participants relate to one another, Sec.~\ref{sec:tactic} further interprets $\mathcal{S}_t$ into a relational traffic context for tactical decision making. At each decision round, the policy outputs
\begin{equation}
    \mathcal{P}_t=(m,\theta_m),
\end{equation}
where $m\in\{\texttt{push-away},\texttt{phantom-braking},\texttt{hold}\}$ denotes the selected physical attack primitive and $\theta_m$ contains its mode-specific continuous parameters. Specifically, $\theta_m=(\Delta d,\rho,o)$ for \texttt{push-away} and $\theta_m=(w,\tau)$ for \texttt{phantom-braking}, as defined in Sec.~\ref{sec:primitives}. A round corresponds to one decision--execution--replanning cycle of the policy in Sec.~\ref{sec:tactic}, and each trial is limited to at most $R=3$ such rounds.

\subsection{Complementary Physical Attack Primitives}
\label{sec:primitives}

\paragraph*{Pointcloud push-away}
Consider a target $T$ following a lead vehicle $A1$, with another vehicle $A2$ behind $T$ in the same lane. The attacker shifts the perceived range of $A1$, causing $T$ to perceive a larger leading gap while the true $T$--$A1$ distance contracts. This primitive, denoted $\texttt{push-away}$, is parameterized by the induced range displacement $\Delta d$, separation ramp $\rho$, and ignition onset $o$.

\paragraph*{Phantom-obstacle braking}
The second primitive injects a virtual obstacle at distance $w$ ahead of $T$. When its perceived TTC falls below the full-braking threshold, $T$ performs emergency braking and the trailing vehicle $A2$ may become the collision agent. This primitive, denoted $\texttt{phantom-braking}$, is parameterized by phantom wall placement distance $w$ and emission duration $\tau$.

\paragraph*{Hold}
The third primitive \texttt{hold} means keeping the previous primitive $\mathcal{P}_{t-1}$. This primitive avoids frequent policy changes during attacking.

The three primitives are complementary because they exploit different traffic relationships. $\texttt{push-away}$ depends primarily on the $T$--$A1$ leader--follower geometry, whereas braking exploits the $A2$--$T$ relationship and the target's deceleration. This complementarity motivates scene-aware tactical selection rather than a fixed attack rule.

\subsection{Physical and Temporal Feasibility}
\label{sec:feasibility}

The available action space is further bounded by physics-informed, empirically calibrated constraints. These constraints determine whether a candidate primitive can operate under the current scene, leaving the scene-aware policy in Sec.~\ref{sec:tactic} to choose among feasible alternatives.

\paragraph*{Push-away feasibility}
The induced displacement must be large enough to produce the required closing behavior while remaining inside roadside sensing coverage:
$
\Delta d_{\min}=11~\mathrm{m},\,
\Delta d_{\max}(o)=50-o~\mathrm{m}.
$
The displacement must also develop before the target reaches the lead vehicle while remaining below the measured consistency threshold:
$
\frac{\Delta d}{t_{\mathrm{arr}}}
\leq \rho \leq
3.0~\mathrm{m/s},
$
where $t_{\mathrm{arr}}$ denotes the remaining arrival time.

\paragraph*{Braking feasibility}
For the braking primitive, the injected obstacle must drive perceived TTC below the 1.6\,s full-braking threshold. We therefore place the virtual wall according to
$
w(v)=1.5v,\,
w\in[5,15]~\mathrm{m},
$
where $v$ is the measured target speed. The emission duration is calibrated to cover both the target's braking response and the follower's subsequent approach.

\paragraph*{Temporal feasibility}
Feasibility depends jointly on traffic geometry and execution time. Waiting can enlarge or shrink relevant inter-vehicle gaps, consume the remaining push-away displacement envelope, or change whether a trailing vehicle can reach the target after braking. Attack timing is therefore determined from the continuously measured spacing, TTC, and vehicle motion rather than by a fixed schedule. Sec.~\ref{sec:tactic} describes how \textsc{TACTIC} combines these feasibility conditions with relational scene understanding for state-adaptive tactical planning.

\begin{figure}[t]
\centering
\includegraphics[width=0.5\textwidth]{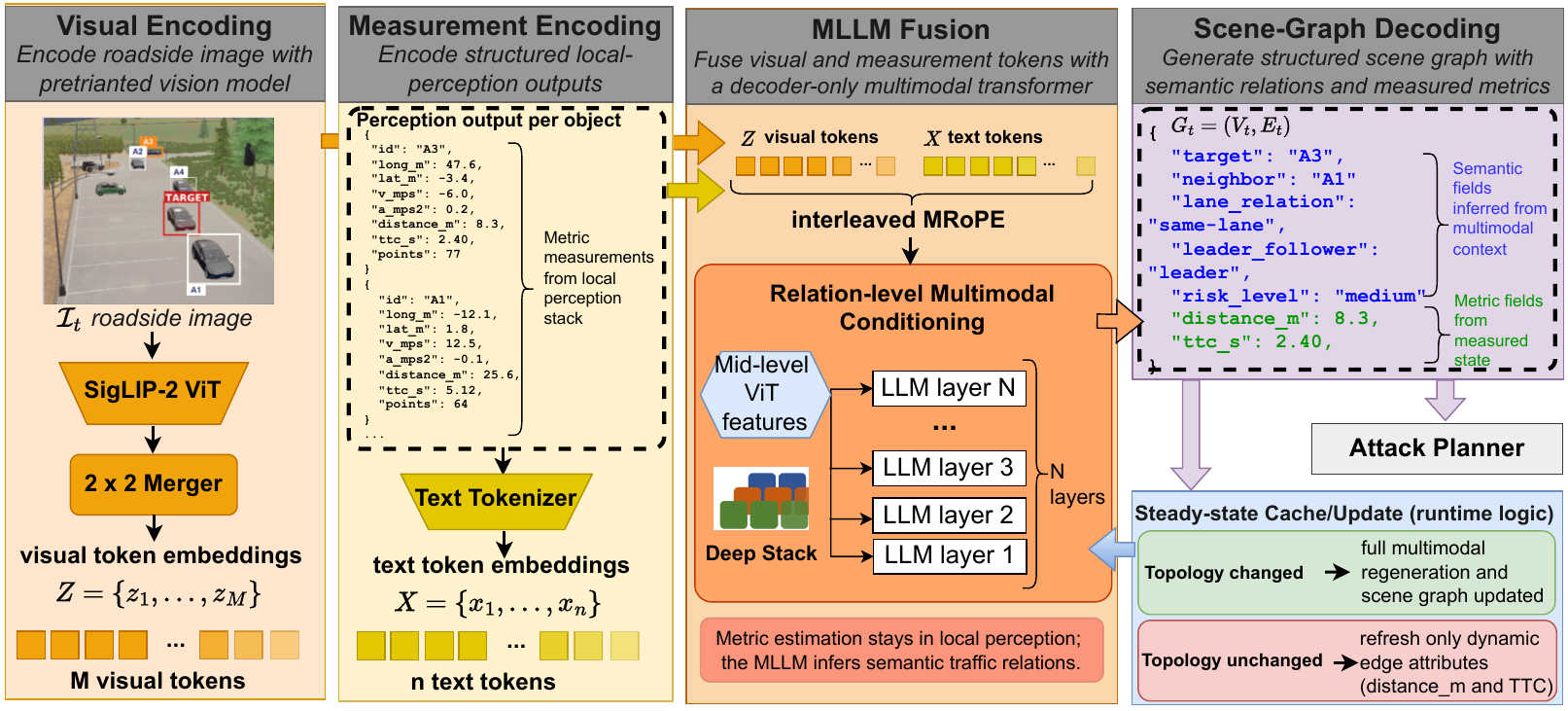}
\caption{Multimodal scene-graph generation in \textsc{TACTIC}. Roadside imagery is encoded into visual tokens, while tracked vehicle states are serialized as text tokens. The MLLM jointly conditions on both modalities to infer semantic traffic relations and decodes them as a structured scene graph. When the topology remains unchanged, the steady-state manager updates only metric edge attributes without full multimodal regeneration.}
\vspace{-10pt}
\label{fig:mllm}
\end{figure}

\section{\textsc{TACTIC}: Scene-Aware LLM Tactical Planning}
\label{sec:tactic}

\textsc{TACTIC} operationalizes the scene-dependent attack space defined in Sec.~\ref{sec:formulation}. Its design separates three responsibilities: local perception measures the traffic state, a multimodal LLM performs relational scene understanding and tactical selection, and deterministic modules enforce the physical constraints in Sec.~\ref{sec:feasibility} and execute the selected primitive. Because perception operates much faster than LLM inference, reasoning and execution proceed asynchronously, with a high-rate local monitor triggering re-planning when the cached scene becomes invalid.

\subsection{Multimodal Semantic Scene Reasoning}
\label{sec:reasoning}

The perception front end operates at the LiDAR's native 20\,Hz rate, clustering detections and associating them into persistent tracks. It provides object-level measurements including longitudinal and lateral position, velocity, acceleration, spacing, and TTC. These measurements accurately describe individual vehicle states but do not explicitly encode their semantic relationships.

To recover this relational context, each MLLM query combines the structured vehicle states with the corresponding roadside camera frame. The image provides complementary road semantics, such as lane structure and driving direction, while the structured measurements retain precise metric information. From these inputs, the MLLM constructs a semantic scene graph
\begin{equation}
G_t=(V_t,E_t),
\end{equation}
where nodes $V_t$ represent tracked traffic participants and edges $E_t$ encode relations such as same-lane leader--follower and adjacent- or opposite-lane interactions. Edge attributes retain metric quantities supplied by the perception stack, including spacing, relative motion, and TTC. This division keeps physical-state estimation outside the MLLM while assigning it the higher-level task of interpreting how traffic participants relate to one another \cite{johnson2015scenegraph,xu2017scenegraph}.

\paragraph*{MLLM scene-graph generator}
Figure~\ref{fig:mllm} illustrates the multimodal generation process. The roadside image $I_t$ is patchified and encoded by a SigLIP-2 vision transformer, followed by a lightweight merger that compresses each $2\times2$ block of patch embeddings into a visual token sequence,
\begin{equation}
Z=\mathrm{Merger}\!\left(\mathrm{ViT}(I_t)\right)=\{z_1,\ldots,z_M\}.
\end{equation}
The tracked vehicle states, output schema, and cached topology are serialized into text tokens $X=\{x_1,\ldots,x_n\}$. Visual and text tokens are jointly processed by the multimodal decoder using interleaved MRoPE positions, while mid-level visual features are injected into early decoder layers through DeepStack. This allows relational fields to condition jointly on visual road context and precise object-state measurements.

The resulting graph is autoregressively decoded into a constrained JSON representation. Letting $y_1,\ldots,y_K$ denote its serialized output tokens,
\begin{equation}
p(G_t)=\prod_{k=1}^{K}p\!\left(y_k\mid y_{<k},Z,X\right).
\end{equation}
Semantic fields, such as lane association and leader--follower relations, are inferred from the multimodal context, whereas metric fields retain measurements supplied by the local perception stack. This separation prevents the MLLM from replacing low-level state estimation while allowing it to resolve traffic relationships needed for tactical planning.

Scene-graph generation runs asynchronously following the generate-while-execute principle of TypeFly \cite{chen2025typefly}. The initial query performs full multimodal scene interpretation. During steady state, subsequent queries reuse the cached graph and recent object states. If the vehicle set and relational topology remain unchanged, the $\Delta$ path refreshes only dynamic edge attributes such as spacing and TTC; otherwise, a vehicle-set consistency check triggers full multimodal regeneration. The updated graph atomically replaces the cached topology once generation completes.

\subsection{Physics-Grounded Tactical Policy}
\label{sec:policy}

Given the semantic graph $G_t$, \textsc{TACTIC} determines which physically feasible primitive best matches the current traffic configuration and how its continuous parameters should be configured. The key design principle is to separate \emph{consequence}, \emph{feasibility}, and \emph{tactical selection} rather than asking the LLM to infer all three from scratch.

\paragraph*{H-group measure}
Before each tactical decision, both primitives are scored using an offline-calibrated h-group measure. For the rear-end primitive, normalized gap, acceleration, and TTC terms use weights 0.15/0.70/0.15, with attack duration introduced at a second combination level with weight 0.10. For braking, acceleration and TTC use weights 0.85/0.15, again with a 0.10 duration term.
Gap is omitted from the braking score because braking severity is driven by the target's deceleration rather than a monotonic function of inter-vehicle distance. Grid-search calibration yields mean attack-to-benign separation margins of $96.5\times$ and $13.6\times$ for the rear-end and braking measures, respectively. The score is defined from the victim's perspective: larger values indicate more severe consequences and do not encode attacker cost or feasibility.

\paragraph*{Feasibility grounding}
A high h-group score does not imply that the corresponding primitive is physically executable. \textsc{TACTIC} therefore supplies the LLM with the measured traffic state together with the mode-specific feasibility bounds derived in Sec.~\ref{sec:feasibility}. For push-away, these include the minimum and coverage-limited displacement,$11 \leq \Delta d \leq 50-o,$ and the admissible ramp band, $\Delta d/t_{\mathrm{arr}} \leq \rho \leq 3.0~\mathrm{m/s}.$

For braking, the wall placement follows the measured target speed, $w(v)=1.5v$ within $[5,15]$\,m, while the minimum emission duration depends on the current follower gap. These values are recomputed from the current scene and provided numerically at every tactical decision.

The LLM therefore reasons over three complementary inputs: relational scene structure, potential consequence, and remaining feasibility margin. A lower-h-group primitive may be selected when the traffic geometry makes it substantially more feasible. Before execution, deterministic code projects the generated parameters back into the admissible set $\Phi(\mathcal{S}_t)$, ensuring that physical bounds are system constraints rather than prompt suggestions. In short, measured states and calibrated constraints determine \emph{what can be executed}, while the LLM determines
\emph{which tactic best matches the scene}.

\paragraph*{Tactical policy} At round $t$, the LLM produces $\mathcal{P}_t=(m,\theta_m)$, where $m$ and $\theta_m$ are defined in Sec.~\ref{sec:attack_obj}. The full policy jointly selects the primitive and its parameters and additionally returns an estimated success likelihood and a short rationale used for logging and analysis. Execution may continue for at most $R=3$ rounds, allowing the tactic to be revised when the previous round fails or the traffic relationship changes.

Attack \emph{timing} is handled separately from tactical selection. The high-rate controller continuously evaluates spacing, TTC, and vehicle motion against the feasible spatio-temporal region defined in Sec.~\ref{sec:feasibility}. It determines whether the cached policy remains executable, should be launched, or must be suspended and reconsidered. The LLM therefore selects and configures the tactic, but does not replace high-rate temporal monitoring.

\paragraph*{Decision sources} To isolate the contribution of LLM tactical reasoning, we compare four decision sources in Sec.~\ref{sec:experiments}. \emph{llm\_policy} jointly selects the primitive and its mode-specific continuous parameters, returning the primitive choice $m$ together with its continuous parameters ($(\Delta d,\rho,o)$ for push-away, $(w,\tau)$ for the phantom wall), with a success estimate and rationale; \emph{llm} selects only the primitive in a single shot without the policy layer, with the continuous parameters fixed at system defaults; \emph{rule} always selects push-away and draws its parameters at random once per trial, exempt from the analytic floors, and reads nothing; and \emph{random} commits to one uniformly sampled tactic and parameterization for the trial.  The random group is not re-sampled after each round, avoiding multiple independent chances that would confound tactical quality.

\subsection{Asynchronous Scene Monitoring and Replanning}
\label{sec:monitoring}

LLM inference takes seconds, whereas the traffic state evolves at the perception rate. \textsc{TACTIC} therefore decouples high-rate scene monitoring from lower-rate semantic and tactical reasoning. While a cached policy is valid, execution continues concurrently with generation of the next scene interpretation. A lightweight local alerter compares the 20\,Hz tracker state against the cached graph on every frame.

A scene mutation is declared when a vehicle appears or disappears, an inter-vehicle gap changes by more than 8\,m or 40\%, or the inferred road context changes. Because ordinary car-following gaps evolve gradually, these thresholds target structural scene changes rather than frame-level noise. A minimum 0.5\,s replanning interval prevents repeated alerts from saturating the LLM scheduler.

When a mutation invalidates the current tactical assumptions, the alerter suspends execution and triggers a scene update. If the vehicle set and relational structure remain valid, the lightweight $\Delta$ path updates the graph; otherwise, a full multimodal regeneration is requested. Tactical planning then resumes from the updated graph and feasibility state. The alerter only detects and localizes scene changes---semantic interpretation and policy regeneration remain the responsibility of the LLM.

This architecture allows perception to react at sensor rate without requiring the LLM to operate at 20\,Hz. 
The resulting latency and scene-change response are evaluated in Sec.~\ref{sec:experiments}.

\section{Experimental Evaluation}
\label{sec:experiments}

We evaluate \textsc{TACTIC} along three dimensions: (i) whether the physical constraints in Sec.~\ref{sec:feasibility} correctly characterize feasible operating regions; (ii) whether LLM-based tactical planning improves attack effectiveness and efficiency over fixed and random policies; and (iii) how multimodal scene reasoning and asynchronous replanning affect performance under changing traffic conditions. We first establish the experimental setting and physical operating bounds, then evaluate tactical decision quality, and finally isolate the contribution of individual system components.

\subsection{Experimental Setup}
\label{sec:setup}

\begin{figure}[t]
\centering
\includegraphics[width=0.49\textwidth]{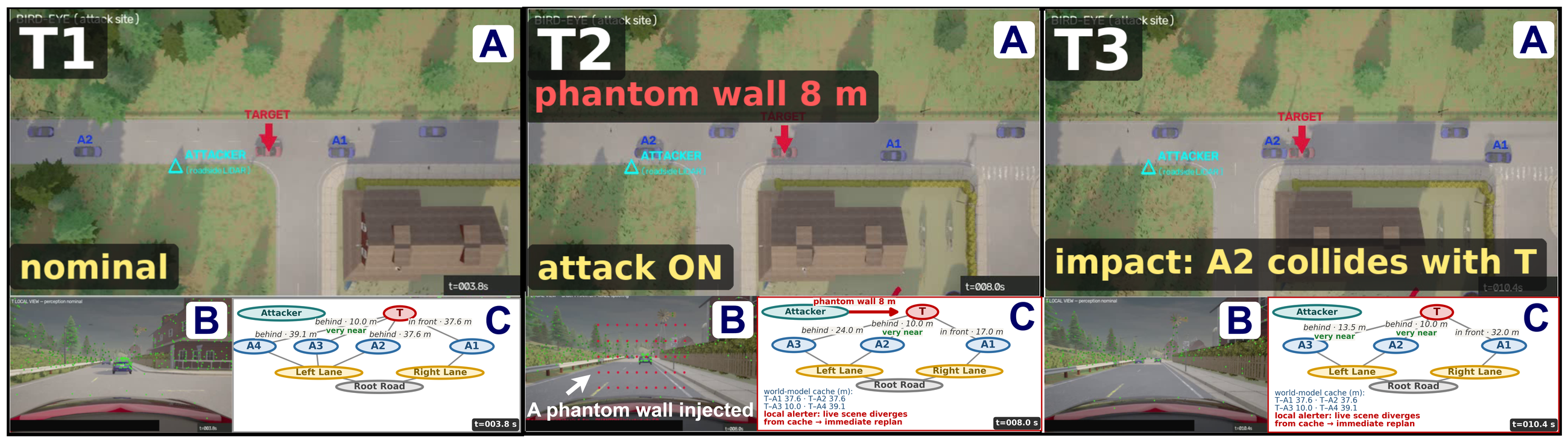}
\caption{Example phantom-obstacle braking attack. At $T_1$ the system monitors a nominal scene; execution begins at $T_2=8.0$\,s and collision occurs at $T_3=10.4$\,s. Windows A--C show the driving scene, target perception, and scene graph.}
\vspace{-10pt}
\label{fig:brake}
\end{figure}

\begin{figure*}[t]
\centering
\includegraphics[width=\textwidth]{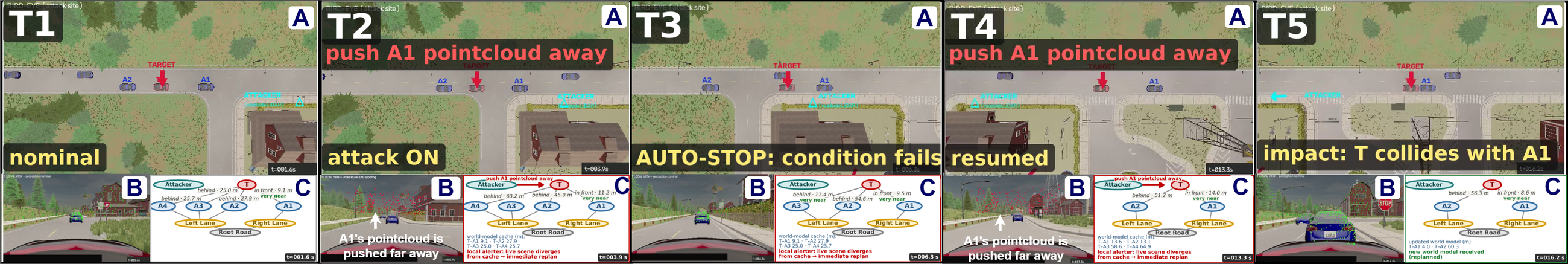}
\caption{Example scene change and replanning. Execution is reconsidered at $T_2=3.9$\,s and $T_4=13.3$\,s as traffic evolves, ultimately causing a collision between $T$ and lead vehicle $A1$.}
\label{fig:scene_change}
\end{figure*}

We implement \textsc{TACTIC} in CARLA 0.9.14 \cite{dosovitskiy2017carla} using Town07 with synchronous simulation at 20\,Hz, as illustrated in Figs.~\ref{fig:brake} and~\ref{fig:scene_change}. Each trial contains the target $T$, a lead vehicle $A1$, a follower $A2$, and two additional traffic participants $A3/A4$. To isolate dynamic vehicle interactions from occlusion, we target the lane closest to the roadside perception stack. The attacker uses pole-mounted LiDAR and an RGB camera in a setting similar to MVS \cite{sato2025realism}, with an effective observation range of approximately 50\,m. The longitudinal vehicles cruise at 6.0\,m/s nominally (closed-loop $\approx5.5$\,m/s), maintaining approximately stationary gaps without attack. Qwen \cite{qwen3vl2025} serves as the primary MLLM backend.

\paragraph*{Monte Carlo protocol}
The initial $T$--$A1$ gap is sampled uniformly from 18--28\,m, spanning the measured push-away feasibility transition. Each configuration contains 20 trials with distinct seeds and at most three attack rounds. Success requires a physical collision, defined as a center-to-center distance below 5.0\,m with residual impact speed of at least 1.5\,m/s. We additionally run 20 attack-free control trials for 30\,s and report Wilson 95\% confidence intervals and Fisher's exact test. Relevant secondary metrics are $t_{90}$, the time to reach 90\% of commanded displacement; $d{<}7$, trials with minimum $T$--$A1$ distance below 7\,m; $R{<}1$, displacement-tracking RMSE below 1\,m; emission Gini $G$; and dose efficiency $\eta$. Together, these metrics capture not only collision success but also physical execution quality and attack efficiency.

\subsection{Feasibility Characterization and Attack Effectiveness}
\label{sec:effectiveness}

We first validate the physical constraints used to ground tactical decisions. These experiments establish the operating regions within which the LLM is allowed to reason, after which we evaluate whether scene-aware policy selection improves attack effectiveness.

\paragraph*{Push-away feasibility}
A locked-displacement sweep reveals a sharp transition: $\Delta d=5$ and 10\,m succeed in 0/20 and 5/20 trials, whereas 15 and 20\,m both succeed in 20/20 (Table~\ref{tab:dose}). A finer sweep gives 1/20 at 8\,m, 0/20 at 9\,m, 5/20 at 10\,m, 12/20 at 10.5\,m, and 20/20 at 11--12\,m, motivating the sufficient floor $\Delta d_{\min}=11$\,m. The upper bound follows sensing coverage, $\Delta d_{\max}=50-o$; beyond the sufficient floor, larger displacement mainly increases exposure rather than improving success.

\begin{figure}[t]
\centering
\subfloat[\label{fig:fig3}]{\vspace{-8pt}\includegraphics[width=0.47\columnwidth]{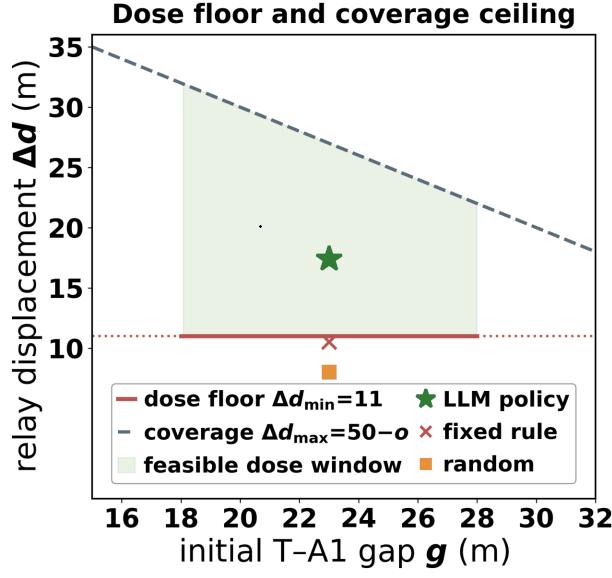}}
\hfill
\subfloat[\label{fig:fig4}]{\vspace{-8pt}\includegraphics[width=0.47\columnwidth]{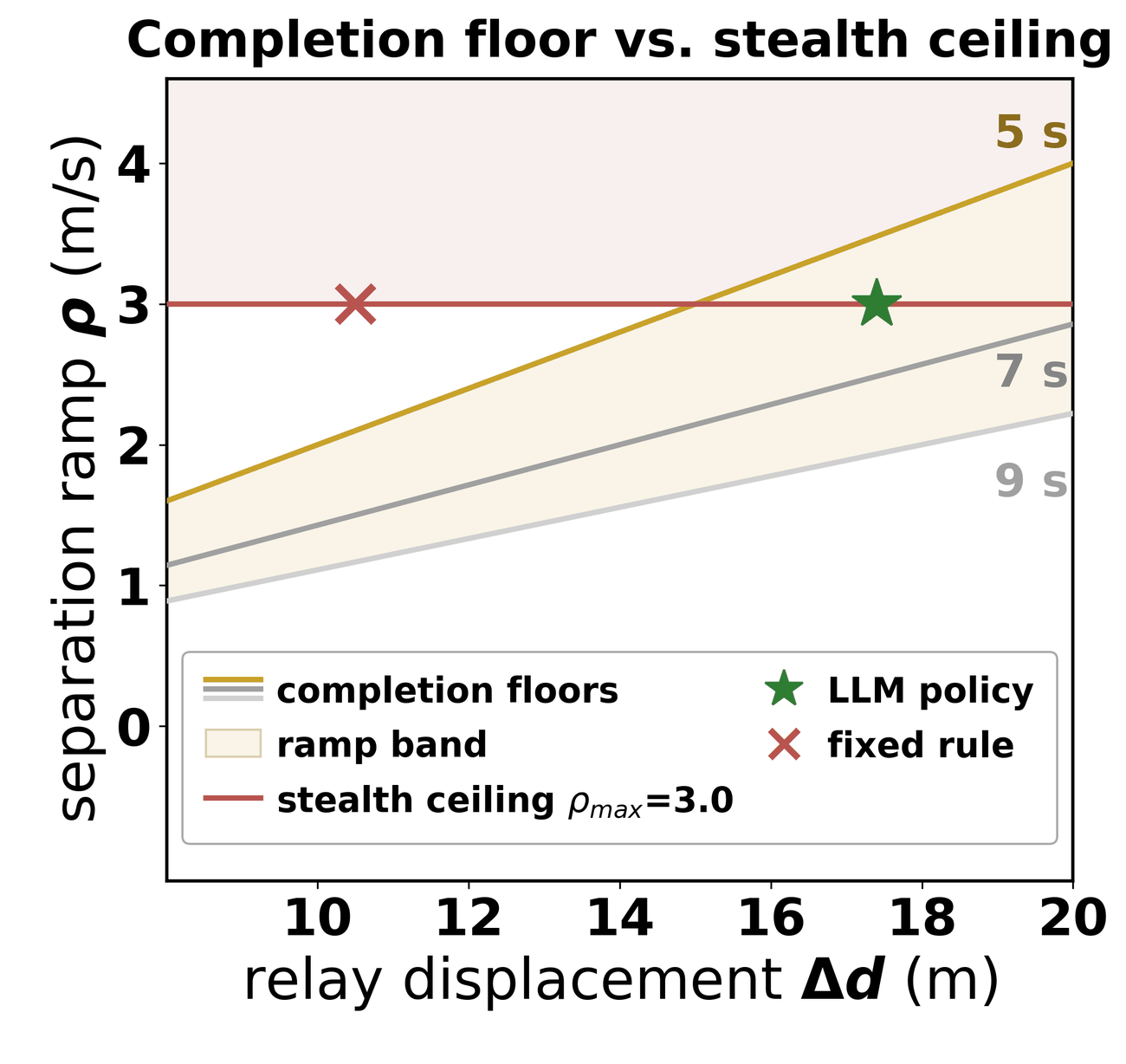}}
\vspace{-5pt}
\caption{Measured push-away feasibility bounds. (a) Displacement range and policy operating points. (b) Ramp rates bounded by completion and the $\rho_{\max}=3.0$\,m/s consistency limit.}
\vspace{-15pt}
\label{fig:pushaway_bounds}
\end{figure}

The ramp must also complete before contact while remaining below the tracking-consistency limit. Instantaneous steps are detected in 10/10 trials and ramps above 3.0\,m/s in 3/3, whereas no tested ramp at or below 3.0\,m/s is flagged (0/10). Together with $\rho\geq\Delta d/t_{\mathrm{arr}}$, this defines the feasible band in Fig.~\ref{fig:pushaway_bounds}. The full policy operates at $\Delta d\approx17.4$\,m and $\rho\approx2.97$\,m/s, placing its typical operating point safely inside this region.

\paragraph*{Phantom-braking feasibility}
We next characterize the complementary braking primitive. A locked sweep over emission duration at a fixed 13--15\,m $A2$--$T$ gap shows no clear duration threshold: success remains high from 2--10\,s. Thus, within this tested range, the binding factor is the follower gap rather than emission duration. The virtual wall follows $w(v)=1.5v$, keeping perceived TTC below the 1.6\,s full-braking threshold. These results confirm that the two primitives are limited by different aspects of scene geometry and therefore occupy distinct state-dependent feasibility regions.

\begin{table}[t]
\centering
\caption{Locked-displacement response of the \texttt{push-away} primitive (20 trials per level; ramp 3.0\,m/s).}
\label{tab:dose}
\footnotesize
\setlength{\tabcolsep}{3.2pt}
\begin{tabular}{ccccc|ccccc}
\toprule
$\Delta d$/m & Succ. & $d{<}7$ & $R{<}1$ & $t_{90}$/s &
$\Delta d$/m & Succ. & $d{<}7$ & $R{<}1$ & $t_{90}$/s \\
\midrule
5  & 0/20  & 0/20  & 0/20  & 5.83 & 15 & 20/20 & 20/20 & 18/20 & 4.87 \\
10 & 5/20  & 20/20 & 9/20  & 4.30 & 20 & 20/20 & 20/20 & 6/20  & 6.20 \\
\bottomrule
\end{tabular}
\vspace{-10pt}
\end{table}

\begin{figure}[t]
\centering
\subfloat[\label{fig:dose_a}]{\vspace{-8pt}\includegraphics[width=0.49\columnwidth]{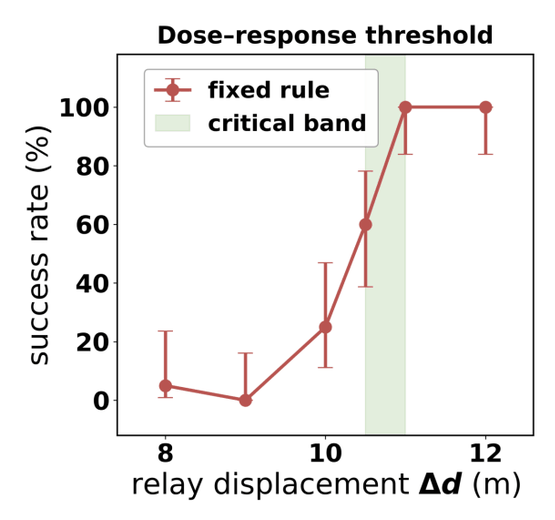}}
\hfill
\subfloat[\label{fig:dose_b}]{\vspace{-8pt}\includegraphics[width=0.49\columnwidth]{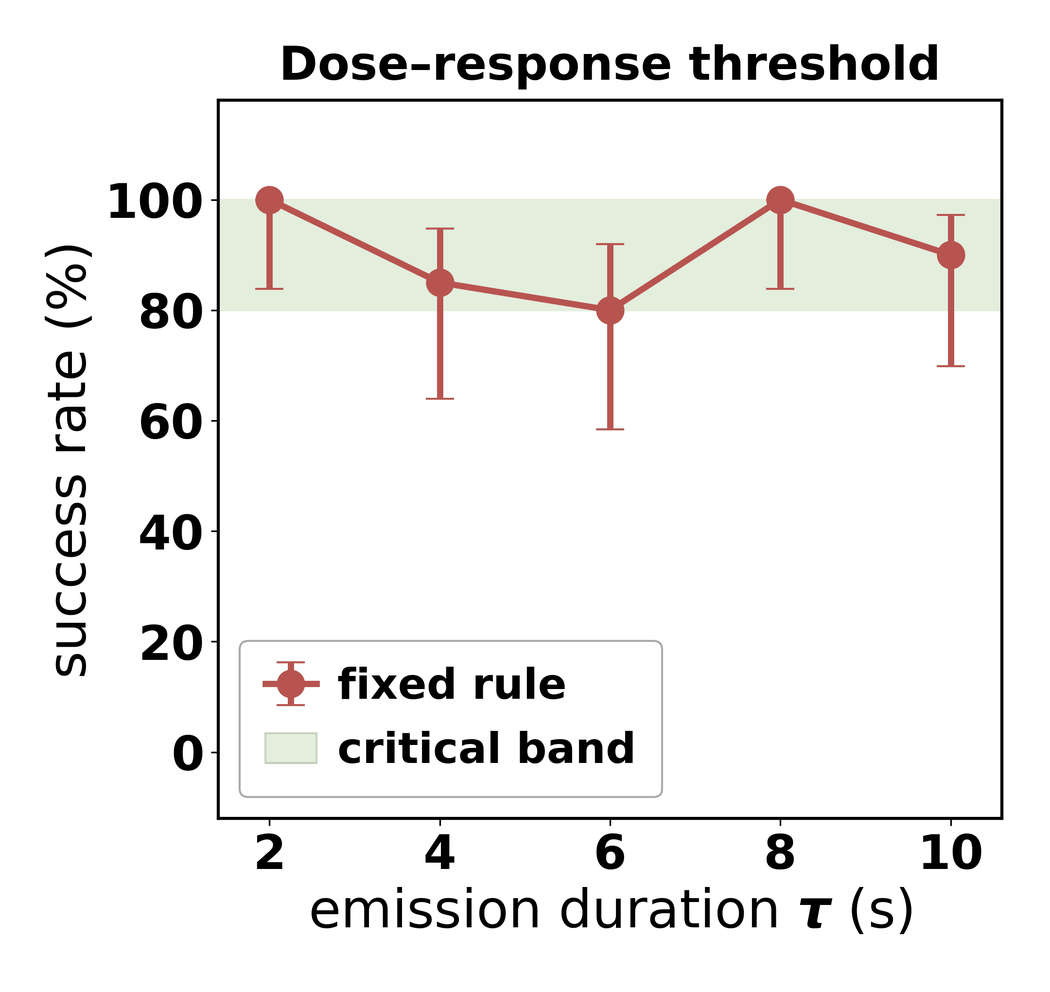}}
\vspace{-5pt}
\caption{Dose--response of the two primitives (20 trials per level; 95\% Wilson CI). (a) \texttt{push-away} success versus relay displacement. (b) \texttt{phantom-braking} success versus emission duration at a fixed 13--15\,m gap.}
\vspace{-20pt}
\label{fig:dose}
\end{figure}

\paragraph*{Dose response}
Figure~\ref{fig:dose} summarizes the resulting dose--response behavior of both primitives. Push-away success rises sharply around 10.5--11\,m and saturates thereafter, validating the 11\,m operating floor. In contrast, phantom-braking remains at or above 80\% success (16/20--20/20) across 2--10\,s emission durations at a fixed 13--15\,m gap, further indicating that follower geometry rather than duration is the dominant constraint. The contrasting response profiles motivate selecting the primitive according to the current traffic state rather than applying one fixed policy.

\paragraph*{Decision-source comparison}
Having established the feasible action space, we next evaluate how different decision mechanisms use it. We compare four decision sources under the hybrid implementation, where both \texttt{push-away} and \texttt{phantom-braking} are available: the full LLM policy, restricted LLM, fixed rule, and committed random policy defined in Sec.~\ref{sec:policy}. The full policy achieves 20/20 successes, significantly exceeding the fixed rule's 7/20 and random policy's 12/20 (Fisher $p=1.3\times10^{-5}$ and $p=0.0033$; Table~\ref{tab:effectiveness}). The fixed rule always chooses push-away, leaving success dependent on sampled geometry; all 13 failures are AEB-arrested near-misses. Random selection can choose either primitive but cannot jointly adapt its parameters to the observed scene.

The restricted LLM achieves 15/20, significantly below the full policy ($p=0.047$), indicating that joint mode and parameter selection improves over mode selection alone. The full policy also has the lowest duty cycle (0.39) and emission dose (3.9), showing that adaptive parameterization improves both effectiveness and signal efficiency. Overall, these results support the central design choice of coupling scene-level primitive selection with continuous parameter adaptation.

\begin{table*}[t]
\centering
\caption{Decision-source comparison under the hybrid implementation (20 trials per group).}
\label{tab:effectiveness}
\footnotesize
\setlength{\tabcolsep}{5.0pt}
\begin{tabular}{llcccccccc}
\toprule
Decision source & Attack decision & Succ.\ [95\% CI] & $t_{90}$/s & $G$ & $\eta$ & LLM calls & Duty & Dose & Fisher $p$ \\
\midrule
LLM policy & LLM-selected mode and $(\Delta d,\rho,o,w,\tau)$ & 20/20 [84,100] & 3.6 & 0.61 & 0.26 & 2.1 & 0.39 & 3.9 & (ref) \\
LLM restricted & LLM-selected mode, default parameters & 15/20 [53,89] & 3.6 & 0.35 & 0.08 & 2.6 & 0.65 & 9.7 & $4.7\times10^{-2}$ \\
Fixed rule & fixed push-away, per-trial random dose & 7/20 [18,57] & 4.3 & 0.33 & 0.07 & -- & 0.67 & 4.7 & $1.3\times10^{-5}$ \\
Random & committed draw over mode and parameters & 12/20 [39,78] & 4.2 & 0.39 & 0.15 & -- & 0.62 & 4.0 & $3.3\times10^{-3}$ \\
\bottomrule
\end{tabular}
\vspace{-10pt}
\end{table*}

\paragraph*{Case studies}
The quantitative results above are further illustrated by two representative executions. Figure~\ref{fig:brake} shows a \texttt{phantom-braking} episode. At $T_2=8.0$\,s, the $A2$--$T$ geometry becomes feasible; the cached policy is released, an 8\,m phantom wall triggers emergency braking, and $A2$ rear-ends $T$ at $T_3=10.4$\,s. Figure~\ref{fig:scene_change} shows replanning under changing traffic. A \texttt{push-away} policy becomes feasible at $T_2=3.9$\,s, but lead-vehicle acceleration invalidates the cached conditions and suspends execution. When feasibility is restored at $T_4=13.3$\,s, the scene graph and policy are updated and execution resumes, eventually causing a $T$--$A1$ collision. Together, these examples illustrate how relational scene reasoning, primitive selection, and high-rate feasibility monitoring interact during execution.

\paragraph*{Benign control}
As a sanity check, no collision occurs in 20 attack-free trials; the minimum separation is 17.9\,m and TTC never enters the hazard region. The observed collisions therefore arise from attack execution rather than the nominal traffic dynamics.

\subsection{Ablation and Runtime Analysis}
\label{sec:ablation}

We finally isolate the contributions of multimodal perception, execution timing, and asynchronous scene updates. These experiments evaluate the components that keep the tactical policy responsive as traffic conditions evolve.

\paragraph*{Perception input}
Removing either input channel significantly reduces success (Table~\ref{tab:awareness}): physical-only input reaches 13/20 ($p=0.0083$) and image-only input 15/20 ($p=0.047$), versus 20/20 with joint input. The two ablations are not significantly different from each other ($p=0.73$), indicating that structured measurements and visual context provide complementary information. The joint representation therefore improves performance by combining precise metric state with semantic road and traffic context.

\begin{table*}[t]
\centering
\vspace{+10pt}
\caption{Ablation and runtime results.}
\label{tab:ablation_summary}
\footnotesize

\subfloat[Perception-input ablation.\label{tab:awareness}]{
\resizebox{0.42\textwidth}{!}{
\begin{tabular}{@{}lccccc@{}}
\toprule
Input & Succ.\ [95\% CI] & Duty & Dose & Time/s & Fisher $p$ \\
\midrule
Physical+image & 20/20 [84,100] & 0.39 & 3.9 & 34.5 & (ref) \\
Physical only & 13/20 [43,82] & 0.49 & 3.2 & 37.8 & $8.3{\times}10^{-3}$ \\
Image only & 15/20 [53,89] & 0.96 & 10.0 & 20.6 & $4.7{\times}10^{-2}$ \\
\bottomrule
\end{tabular}
}}
\hspace{0.015\textwidth}
\subfloat[Scene-mutation latency.\label{tab:latency}]{
\resizebox{0.20\textwidth}{!}{
\begin{tabular}{@{}lc@{}}
\toprule
Configuration & Latency/s \\
\midrule
Synchronous & 7.4 \\
Asynchronous & 3.7 \\
Async. + $\Delta$ refresh & 2.0 \\
\bottomrule
\end{tabular}
}}
\hspace{0.015\textwidth}
\subfloat[Attack-timing ablation.\label{tab:timing}]{
\resizebox{0.32\textwidth}{!}{
\begin{tabular}{@{}lccc@{}}
\toprule
Timing & Succ.\ [95\% CI] & Time/s & Fisher $p$ \\
\midrule
Immediate & 20/20 [84,100] & 34.5 & (ref) \\
Wait for window & 9/20 [26,66] & 43.7 & $1.5{\times}10^{-4}$ \\
\bottomrule
\end{tabular}
}}
\vspace{-15pt}
\end{table*}

\paragraph*{Attack timing}
We next test whether deliberately waiting for a nominally vulnerable state improves execution. Immediate execution achieves 20/20 successes, versus 9/20 when waiting for a predefined vulnerable window (gap $<12$\,m and closing speed $>0.5$\,m/s within 4\,s), with Fisher $p=1.5\times10^{-4}$ (Table~\ref{tab:timing}). Waiting also increases mean trial time from 34.5 to 43.7\,s. Thus, delaying execution can consume the spatio-temporal feasibility envelope before the desired condition is reached; we therefore execute once the current state satisfies the primitive constraints.

\paragraph*{Asynchronous replanning}
Finally, we evaluate whether asynchronous processing reduces the latency introduced by MLLM reasoning. The reasoning channel makes approximately 2.5 LLM calls per trial. During steady traffic, all ten $\Delta$-path probes correctly preserve the existing topology while updating metric edge attributes. Under injected scene mutations, the local alerter suspends execution within 0.05\,s. Synchronous regeneration requires 7.4\,s before execution resumes, asynchronous regeneration reduces this to 3.7\,s, and asynchronous $\Delta$ refresh further reduces it to 2.0\,s when topology remains unchanged (Table~\ref{tab:latency}). Decoupling high-rate monitoring from lower-rate MLLM reasoning therefore substantially reduces scene-update latency while preserving rapid local response.

\paragraph*{Backend generality}
To test whether these results depend on a particular MLLM, we repeat both LLM decision groups using Kimi \cite{team2025kimi}. Qwen \cite{qwen3vl2025} achieves 20/20 and 15/20 successes for the full and restricted policies, respectively, while Kimi achieves 20/20 and 16/20. The differences are not significant (Fisher $p=1.0$), suggesting that the observed advantage is not specific to one MLLM backend.

\section{Conclusion}

This paper presents \textsc{TACTIC}, a scene-dependent framework for physical LiDAR attack planning in dynamic traffic. \textsc{TACTIC} combines multimodal LLM-based relational reasoning with calibrated physical constraints and asynchronous replanning to select feasible attack tactics from evolving scene context. Across 280 CARLA trials in five ablation experiments, the full policy achieves a 100\% collision rate across both primitives, significantly above the fixed-rule (35\%) and random (60\%) baselines. Joint physical-and-image perception sustains 100\%, whereas single-channel ablations drop to 65\% and 75\%. Asynchronous replanning reduces scene-change response latency from 7.4\,s to 2.0\,s. These results demonstrate that scene-aware tactical planning can expose failure modes missed by static attack evaluation. Future work will extend \textsc{TACTIC} to hardware platforms, learned perception stacks, and richer sequence-level tactics.

\bibliographystyle{IEEEtran}
\bibliography{reference}

@inproceedings{jin2023pla,
  title={Pla-lidar: Physical laser attacks against lidar-based 3d object detection in autonomous vehicle},
  author={Jin, Zizhi and Ji, Xiaoyu and Cheng, Yushi and Yang, Bo and Yan, Chen and Xu, Wenyuan},
  booktitle={2023 IEEE Symposium on Security and Privacy (SP)},
  pages={1822--1839},
  year={2023},
  organization={IEEE}
}

@inproceedings{zhang2025ghost,
  title={The ghost navigator: Revisiting the hidden vulnerability of localization in autonomous driving},
  author={Zhang, Junqi and Cheng, Shaoyin and Hu, Linqing and Zhang, Jie and Shi, Chengyu and Han, Xingshuo and Zhang, Tianwei and Cheng, Yueqiang and Zhang, Weiming},
  booktitle={34th USENIX Security Symposium (USENIX Security 25)},
  pages={3979--3998},
  year={2025}
}

@inproceedings{sato2025realism,
  title={On the realism of lidar spoofing attacks against autonomous driving vehicle at high speed and long distance},
  author={Sato, Takami and Suzuki, Ryo and Hayakawa, Yuki and Ikeda, Kazuma and Sako, Ozora and Nagata, Rokuto and Yoshida, Ryo and Chen, Qi Alfred and Yoshioka, Kentaro},
  booktitle={Network and Distributed System Security Symposium (NDSS)},
  year={2025}
}

@inproceedings{cao2023you,
  title={You can't see me: Physical removal attacks on $\{$lidar-based$\}$ autonomous vehicles driving frameworks},
  author={Cao, Yulong and Bhupathiraju, S Hrushikesh and Naghavi, Pirouz and Sugawara, Takeshi and Mao, Z Morley and Rampazzi, Sara},
  booktitle={32nd USENIX security symposium (USENIX Security 23)},
  pages={2993--3010},
  year={2023}
}

@inproceedings{cao2019adversarial,
  title={Adversarial sensor attack on lidar-based perception in autonomous driving},
  author={Cao, Yulong and Xiao, Chaowei and Cyr, Benjamin and Zhou, Yimeng and Park, Won and Rampazzi, Sara and Chen, Qi Alfred and Fu, Kevin and Mao, Z Morley},
  booktitle={Proceedings of the 2019 ACM SIGSAC conference on computer and communications security},
  pages={2267--2281},
  year={2019}
}

@article{zhang2025sok,
  title={SoK: How sensor attacks disrupt autonomous vehicles: An end-to-end analysis, challenges, and missed threats},
  author={Zhang, Qingzhao and Luo, Shaocheng and Mao, Z Morley and Pajic, Miroslav and Reiter, Michael K},
  journal={arXiv preprint arXiv:2509.11120},
  year={2025}
}

@inproceedings{song2025gradient,
  title={Gradient-Based Adversarial Attacks on Deep LiDAR Odometry},
  author={Song, Zhenbo and Chen, Xuanzhu and Zhang, Zhenyuan and Zhang, Kaihao and Lu, Jianfeng and Li, Weiqing},
  booktitle={2025 IEEE International Conference on Robotics and Automation (ICRA)},
  pages={15188--15194},
  year={2025},
  organization={IEEE}
}

@inproceedings{nassi2020phantom,
  title={Phantom of the ADAS: Securing advanced driver-assistance systems from split-second phantom attacks},
  author={Nassi, Ben and Mirsky, Yisroel and Nassi, Dov and Ben-Netanel, Reut and Drokin, Oleg and Elovici, Yuval},
  booktitle={Proceedings of the 2020 ACM SIGSAC Conference on Computer and Communications Security},
  pages={293--308},
  year={2020}
}

@inproceedings{sun2020robust,
  title={Towards robust LiDAR-based perception in autonomous driving: General black-box adversarial sensor attack and countermeasures},
  author={Sun, Jingjing and Cao, Yulong and Chen, Qi Alfred and Mao, Z Morley},
  booktitle={29th USENIX Security Symposium},
  pages={877--894},
  year={2020}
}

@inproceedings{tu2020physically,
  title={Physically realizable adversarial examples for LiDAR object detection},
  author={Tu, Jing and Ren, Mengye and Manivasagam, Sivabalan and Liang, Ming and Yang, Bin and Du, Rui and Cheng, Fangcheng and Urtasun, Raquel},
  booktitle={Proceedings of the IEEE/CVF Conference on Computer Vision and Pattern Recognition (CVPR)},
  pages={13716--13725},
  year={2020}
}

@inproceedings{kim2024hardness,
  title={A systematic study of physical sensor attack hardness},
  author={Kim, Hyoungshick and Bandyopadhyay, Riad and Ozmen, Mehmet Oguz and Celik, Z Berkay and Bianchi, Andrea and Kim, Yousra and Xu, Dongyan},
  booktitle={2024 IEEE Symposium on Security and Privacy (SP)},
  year={2024},
  organization={IEEE}
}

@article{mei2025llmattacker,
  title={LLM-attacker: Enhancing closed-loop adversarial scenario generation for autonomous driving with large language models},
  author={Mei, Yutong and Nie, Tianyu and Sun, Jian and Tian, Yu},
  journal={arXiv preprint arXiv:2501.15850},
  year={2025}
}

@article{chen2025typefly,
  title={TypeFly: Low-latency drone planning with large language models},
  author={Chen, Guojun and Yu, Xiaojing and Ling, Neiwen and Zhong, Lin},
  journal={IEEE Transactions on Mobile Computing},
  volume={24},
  number={9},
  pages={9068--9079},
  year={2025}
}

@article{yao2026seidm,
  title={{SEIDM}: A safe and efficient intelligent driver model for autonomous driving behavior},
  author={Yao, Y. and Luo, S.},
  journal={arXiv preprint arXiv:2605.23915},
  year={2026}
}

@misc{euroncap2024,
  author={{European New Car Assessment Programme (Euro NCAP)}},
  title={Test protocol -- AEB/LSS VRU systems},
  howpublished={v4.5.1, Feb. 2024},
  year={2024}
}

@inproceedings{johnson2015scenegraph,
  title={Image retrieval using scene graphs},
  author={Johnson, Justin and Krishna, Ranjay and Stark, Michael and Li, Li-Jia and Shamma, David A. and Bernstein, Michael and Fei-Fei, Li},
  booktitle={Proceedings of the IEEE Conference on Computer Vision and Pattern Recognition (CVPR)},
  pages={3668--3678},
  year={2015}
}

@inproceedings{xu2017scenegraph,
  title={Scene graph generation by iterative message passing},
  author={Xu, Danfei and Zhu, Yuke and Choy, Christopher B. and Fei-Fei, Li},
  booktitle={Proceedings of the IEEE Conference on Computer Vision and Pattern Recognition (CVPR)},
  pages={5410--5419},
  year={2017}
}

@inproceedings{dosovitskiy2017carla,
  title={CARLA: An open urban driving simulator},
  author={Dosovitskiy, Alexey and Ros, German and Codevilla, Felipe and Lopez, Antonio and Koltun, Vladlen},
  booktitle={Proceedings of the 1st Annual Conference on Robot Learning (CoRL)},
  pages={1--16},
  year={2017}
}

@misc{qwen3vl2025,
  title={Qwen3-VL technical report},
  author={{Qwen Team}},
  howpublished={arXiv preprint arXiv:2511.21631},
  year={2025}
}

@inproceedings{sima2024drivelm,
  title={Drivelm: Driving with graph visual question answering},
  author={Sima, Chonghao and Renz, Katrin and Chitta, Kashyap and Chen, Li and Zhang, Hanxue and Xie, Chengen and Bei{\ss}wenger, Jens and Luo, Ping and Geiger, Andreas and Li, Hongyang},
  booktitle={European conference on computer vision},
  pages={256--274},
  year={2024},
  organization={Springer}
}

@article{mao2023gpt,
  title={Gpt-driver: Learning to drive with gpt},
  author={Mao, Jiageng and Qian, Yuxi and Ye, Junjie and Zhao, Hang and Wang, Yue},
  journal={arXiv preprint arXiv:2310.01415},
  year={2023}
}

@inproceedings{shao2024lmdrive,
  title={Lmdrive: Closed-loop end-to-end driving with large language models},
  author={Shao, Hao and Hu, Yuxuan and Wang, Letian and Song, Guanglu and Waslander, Steven L and Liu, Yu and Li, Hongsheng},
  booktitle={2024 IEEE/CVF Conference on Computer Vision and Pattern Recognition (CVPR)},
  pages={15120--15130},
  year={2024},
  organization={IEEE}
}

@article{kesting2010enhanced,
  title={Enhanced intelligent driver model to access the impact of driving strategies on traffic capacity},
  author={Kesting, Arne and Treiber, Martin and Helbing, Dirk},
  journal={Philosophical Transactions: Mathematical, Physical and Engineering Sciences},
  pages={4585--4605},
  year={2010},
  publisher={JSTOR}
}

@article{team2025kimi,
  title={Kimi k2: Open agentic intelligence},
  author={Team, Kimi and Bai, Yifan and Bao, Yiping and Charles, Y and Chen, Cheng and Chen, Guanduo and Chen, Haiting and Chen, Huarong and Chen, Jiahao and Chen, Ningxin and others},
  journal={arXiv preprint arXiv:2507.20534},
  year={2025}
}

\end{document}